\documentclass[10pt,twocolumn,letterpaper]{article}

\usepackage{iccv}
\usepackage{times}
\usepackage{epsfig}
\usepackage{graphicx}
\usepackage{amsmath}
\usepackage{amssymb}

\usepackage{color}
\usepackage{makecell}
\usepackage{mathrsfs}
\usepackage[dvipsnames]{xcolor}
\usepackage{multirow}
\usepackage{colortbl}

\usepackage[pagebackref=true,breaklinks=true,letterpaper=true,colorlinks,bookmarks=false]{hyperref}
\usepackage{rotating}

\usepackage{array}
\newcolumntype{L}[1]{>{\raggedright\let\newline\\\arraybackslash\hspace{0pt}}m{#1}}
\newcolumntype{C}[1]{>{\centering\let\newline\\\arraybackslash\hspace{0pt}}X{#1}}
\newcolumntype{R}[1]{>{\raggedleft\let\newline\\\arraybackslash\hspace{0pt}}m{#1}}
\newcommand*{\affmark}[1][*]{\textsuperscript{#1}}

\iccvfinalcopy 

\def\iccvPaperID{5417} 

\begin{document}

\title{SPair-71k: A Large-scale Benchmark for Semantic Correspondence}


\author{Juhong Min\affmark[1,2]\hspace{0.8cm}Jongmin Lee\affmark[1,2]\hspace{0.8cm}Jean Ponce\affmark[3,4]\hspace{0.8cm}Minsu Cho\affmark[1,2]\vspace{1.5mm}\\
\affmark[1]POSTECH \hspace{1.5cm} 
\affmark[2]NPRC\footnotemark[1] \hspace{1.5cm}  
\affmark[3]Inria\vspace{3.0mm} \hspace{1.5cm} 
\affmark[4]DI ENS\footnotemark[2] \\
}

\maketitle

\newcommand{\mcho}[1]{\textcolor{magenta}{#1}}
\newcommand{\jmin}[1]{\textcolor{blue}{#1}}
\newcommand{\jmlee}[1]{\textcolor{ForestGreen}{#1}}

\footnotetext[1]{The Neural Processing Research Center, Seoul, Korea}
\footnotetext[2]{D\'epartement d'informatique de l'ENS, ENS, CNRS, PSL University, Paris, France}

\begin{abstract}
Establishing visual correspondences under large intra-class variations, which is often referred to as semantic correspondence or semantic matching, remains a challenging problem in computer vision. 
Despite its significance, however, most of the datasets for semantic correspondence are limited to a small amount of image pairs with similar viewpoints and scales.
In this paper, we present a new large-scale benchmark dataset of semantically paired images, {\normalfont\textbf{SPair-71k}}, which contains 70,958 image pairs with diverse variations in viewpoint and scale. Compared to previous datasets, it is significantly larger in number and contains more accurate and richer annotations. 
We believe this dataset will provide a reliable testbed to study the problem of semantic correspondence and will help to advance research in this area.  We provide the results of recent methods on our new dataset as baselines for further research. Our benchmark is available online at {\small \url{http://cvlab.postech.ac.kr/research/SPair-71k/}}.
\end{abstract}

\section{Motivation}

The problem of semantic correspondence aims at establishing visual correspondences between images depicting different instances of the same object or scene category~\cite{ham2016proposal,liu2016sift}. Unlike other conventional problems of visual correspondence such as stereo matching, optical flow, and wide-baseline matching, 
it inherently involves a variety of intra-class variations, which makes the problem notoriously challenging. 
With growing interest in semantic correspondence, several annotated benchmarks are now available. 
Due to the high expense of ground-truth annotations for semantic correspondence, early benchmarks~\cite{chen_cvpr14, kim2013deformable} only support indirect evaluation using a surrogate evaluation metric rather than direct matching accuracy.    
For example, the Caltech-101 dataset in~\cite{kim2013deformable} provides binary mask annotations of objects of interest for 1,515 pairs of images and the accuracy of mask transfer is evaluated as a rough approximation to that of matching. 
Recently, Ham~\etal~\cite{ham2016proposal,ham2018proposal} and Taniai~\etal~\cite{taniai2016joint} have introduced datasets with ground-truth correspondences. Since then, PF-WILLOW~\cite{ham2016proposal} and PF-PASCAL~\cite{ham2018proposal} have been used for evaluation in many papers. They contain 900 and 1,300 image pairs, respectively, with keypoint annotations for semantic parts. 

\begin{figure}[t]
    \begin{center}
      \includegraphics[width=1.0\linewidth]{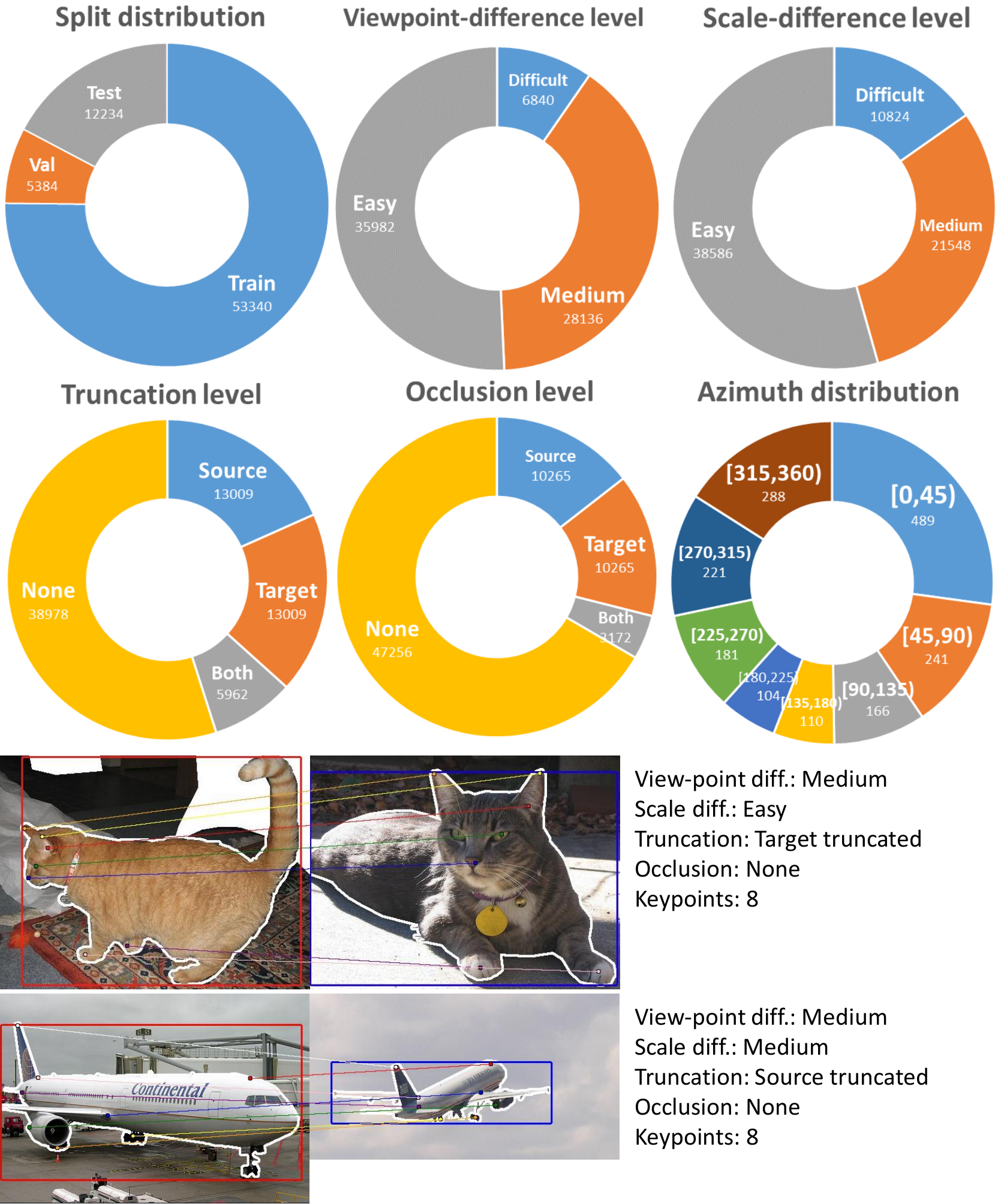}
    \end{center}
    \vspace{-3.0mm}
    \caption{SPair-71k data statistics and example pairs with its annotations. Best viewed in electronic form.}
    \label{fig:dataset_summary_example}
    \vspace{-2.0mm}
\end{figure}

\begin{table*}
    \begin{center}
        \scalebox{0.8}{
        \centering
        \begin{tabular}{c|ccc|ccc|cccc|cccc}
            \hline
            \multirow{2}{*}{Type} & \multicolumn{3}{c|}{View-point diff.} & \multicolumn{3}{c|}{Scale diff.} & \multicolumn{4}{c|}{Truncation diff.} & \multicolumn{4}{c}{Occlusion diff.} \\
             & easy  & medi & hard &  easy  & medi & hard &  none  & src & tgt &  both  &  none  & src & tgt &  both  \\
            \hline
Train  &  26,466 &  21,646 &   5,228 &  29,248 &  16,184 &   7,908 &  29,184 &   9,796 &   9,796 &   4,564 &  35,330 &   7,737 &   7,737 &   2,536 \\
Val  &   2,862 &   2,016 &    506 &   2,880 &   1,570 &    934 &   2,744 &   1,047 &   1,047 &    546 &   3,760 &    722 &    722 &    180 \\
Test &   6,654 &   4,474 &   1,106 &   6,458 &   3,794 &   1,982 &   7,050 &   2,166 &   2,166 &    852 &   8,166 &   1,806 &   1,806 &    456 \\
            \hline
All  &  35,982 &  28,136 &   6,840 &  38,586 &  21,548 &  10,824 &  38,978 &  13,009 &  13,009 &   5,962 &  47,256 &  10,265 &  10,265 &   3,172 \\
            \hline
        \end{tabular}
        }
        \vspace{+1.0mm}
        \caption{Distribution of SPair-71k in terms of difficulty labels.}
        \label{tab:large_split_additional}
        \vspace{-8.0mm}
    \end{center}
\end{table*}

\begin{table}[t]
    \begin{center}
    \scalebox{0.85}{
    \begin{tabular}{c|cccc}
        \hline
        Category & Train &  Val & Test &  All \\
        \hline
      aeroplane &     2,924 &      304 &      690 &     3,918 \\
        bicycle &     2,946 &      334 &      650 &     3,930 \\
           bird &     3,080 &      272 &      702 &     4,054 \\
           boat &     2,936 &      302 &      702 &     3,940 \\
         bottle &     2,448 &      374 &      870 &     3,692 \\
            bus &     2,708 &      250 &      644 &     3,602 \\
            car &     2,960 &      274 &      564 &     3,798 \\
            cat &     3,306 &      272 &      600 &     4,178 \\
          chair &     3,060 &      306 &      646 &     4,012 \\
            cow &     3,118 &      272 &      640 &     4,030 \\
            dog &     3,192 &      306 &      600 &     4,098 \\
          horse &     3,136 &      306 &      600 &     4,042 \\
      motorbike &     3,180 &      238 &      702 &     4,120 \\
         person &     3,066 &      306 &      650 &     4,022 \\
    potted plant &     2,446 &      380 &      862 &     3,688 \\
          sheep &     3,140 &      240 &      664 &     4,044 \\
          train &     2,752 &      342 &      756 &     3,850 \\
      tv/monitor &     2,942 &      306 &      692 &     3,940 \\
        \hline
        total &    53,340 &     5,384 &    12,234 &    70,958 \\
        \hline
    \end{tabular}
    }
    \vspace{+1.0mm}
    \caption{Distribution of SPair-71k in terms of category labels.}
    \label{tab:large_split}
    \vspace{-8.0mm}
    \end{center}
\end{table}

All previous datasets, however, have several drawbacks: 
First, the amount of data is not sufficient to train and test a large model.
Second, image pairs do not display much variability in viewpoint, scale, occlusion, and truncation. 
Third, the annotations are often limited to either keypoints or object segmentation masks, which hinders in-depth analysis.
Fourth, the datasets have no clear splits for training, validation, and testing. Due to this, recent evaluations in~\cite{han2017scnet, Rocco18, rocco2018neighbourhood} have been done with different dataset splits of PF-PASCAL. Furthermore, the splits are disjoint in terms of image pairs, but not images: some images are shared between training and testing data. 

To resolve these issues, we introduce a new dataset, {\em SPair-71k}, consisting of total 70,958 pairs of images from PASCAL 3D+~\cite{Xiang2014BeyondPA} and PASCAL VOC 2012~\cite{everingham2015pascal}. The dataset is significantly larger with rich annotations and clearly organized for learning.   
In particular, several types of useful annotations are available: keypoints of semantic parts, object segmentation masks, bounding boxes, view-point, scale, truncation, and occlusion differences for image pairs, etc.
Figure~\ref{fig:dataset_summary_example} shows the dataset statistics in pie chart forms and sample image pairs with their annotations. Our benchmark is available online at {\tt\small \url{http://cvlab.postech.ac.kr/research/SPair-71k/}}.

$^*$This article is extended from section 4 of our recent paper~\cite{min2019hyperpixel} to provide the details of the dataset and more results.


\begin{center}
    \begin{table*}
        \begin{center}
            \scalebox{0.65}{
            \begin{tabular}{c|c|cccccccccccccccccc|c}
            \hline
             \multicolumn{2}{c|}{Methods} & aero & bike & bird & boat & bottle & bus & car & cat & chair & cow & dog & horse & moto & person & plant & sheep & train & tv & All\\
            \hline
            \hline
            \multirow{6}{*}{\shortstack[1]{Authors' \\ original \\ models}}
            & CNNGeo$_\textrm{res101}$~\cite{Rocco17} & 21.3 & 15.1 & 34.6 & 12.8 & 31.2 & 26.3 & 24.0 & 30.6 & 11.6 & 24.3 & 20.4 & 12.2 & 19.7 & 15.6 & 14.3 & 9.6 & 28.5 & 28.8 & 18.1  \\
            & A2Net$_\textrm{res101}$~\cite{paul2018attentive} &  20.8 & 17.1 & 37.4 & 13.9 & 33.6 & \underline{29.4} & \underline{26.5} & 34.9 & 12.0 & 26.5 & 22.5 & 13.3 & 21.3 & 20.0 & 16.9 & 11.5 & 28.9 & 31.6 & 20.1 \\
            & WeakAlign$_\textrm{res101}$~\cite{Rocco18} & 23.4 & 17.0 & 41.6 & 14.6 & 37.6 & \underline{28.1} & \underline{26.6} & 32.6 & 12.6 & 27.9 & 23.0 & 13.6 & 21.3 & 22.2 & 17.9 & 10.9 & \underline{31.5} & 34.8 & 21.1 \\
            & NC-Net$_\textrm{res101}$~\cite{rocco2018neighbourhood} & \underline{24.0} & 16.0 & \underline{45.0} & 13.7 & 35.7 & 25.9 & 19.0 & \underline{50.4} & \underline{14.3} & \underline{32.6} & \underline{27.4} & \underline{19.2} & \underline{21.7} & 20.3 & 20.4 & \underline{13.6}  & \textbf{33.6} & \textbf{40.4} & \underline{26.4} \\
            & HPF$_\mathrm{res50}$~\cite{min2019hyperpixel} & \textbf{25.3} & \underline{18.5} & \underline{47.6} & 14.6 & 37.0 & 22.9 & 18.3 & \underline{51.1} & \underline{16.7} & \underline{31.5} & \underline{30.8} & \underline{19.1} & \underline{23.7} & \underline{23.8} & \textbf{23.5} & \underline{14.4} & 30.8 & \underline{37.2} & \underline{27.2} \\
            & HPF$_\mathrm{res101}$~\cite{min2019hyperpixel} & \underline{25.2} & \textbf{18.9} & \textbf{52.1} & \underline{15.7} & \underline{38.0} & 22.8 & 19.1 & \textbf{52.9} & \textbf{17.9} & \textbf{33.0} & \textbf{32.8} & \textbf{20.6} & \textbf{24.4} & \textbf{27.9} & \underline{21.1} & \textbf{15.9} & \underline{31.5} & 35.6 & \textbf{28.2} \\
            
            
            \hline
            \multirow{4}{*}{\shortstack[1]{SPair-71k \\ finetuned \\  models}}
            & CNNGeo$_\textrm{res101}$~\cite{Rocco17} &  23.4 & 16.7 & 40.2 & 14.3 & 36.4 & 27.7 & 26.0 & 32.7 & 12.7 & 27.4 & 22.8 & 13.7 & 20.9 & 21.0 & 17.5 & 10.2 & 30.8 & 34.1 & 20.6  \\
            & A2Net$_\textrm{res101}$~\cite{paul2018attentive} & 22.6 & \underline{18.5} & 42.0 & \textbf{16.4} & \underline{37.9} & \textbf{30.8} & \underline{26.5} & 35.6 & 13.3 & 29.6 & 24.3 & 16.0 & 21.6 & \underline{22.8} & \underline{20.5} & 13.5 & 31.4 & \underline{36.5} & 22.3 \\
            & WeakAlign$_\textrm{res101}$~\cite{Rocco18} &  22.2 & 17.6 & 41.9 & \underline{15.1} & \textbf{38.1} & 27.4 & \textbf{27.2} & 31.8 & 12.8 & 26.8 & 22.6 & 14.2 & 20.0 & 22.2 & 17.9 & 10.4 & \underline{32.2} & 35.1 & 20.9 \\
            & NC-Net$_\textrm{res101}$~\cite{rocco2018neighbourhood} & 17.9 & 12.2 & 32.1 & 11.7 & 29.0 & 19.9 & 16.1 & 39.2 & 9.9 & 23.9 & 18.8 & 15.7 & 17.4 & 15.9 & 14.8 & 9.6 & 24.2 & 31.1 & 20.1   \\
            \hline
            \end{tabular}}
        \vspace{+2.0mm}
        \caption{\label{tab:hpftable}Per-class PCK ($\alpha_{\text{bbox}}=0.1$) results on SPair-71k dataset. For the authors' original models, the models of~\cite{Rocco17, paul2018attentive} trained on PASCAL-VOC with self-supervision, ~\cite{Rocco18, rocco2018neighbourhood} trained on PF-PASCAL with weak-supervision, and ~\cite{min2019hyperpixel} tuned using validation split of SPair-71k are used for evaluation.
        For SPair-71k-finetuned models, the original models are further finetuned on SPair-71k dataset by ourselves with our best efforts. Numbers in bold indicate the best performance and underlined ones are the second and third best.}
        \end{center}
    \end{table*}
\end{center} 

\begin{center}
    \begin{table*}
        \begin{center}
            \scalebox{0.8}{
            \begin{tabular}{c|c|ccc|ccc|cccc|cccc|c}
            \hline
            \multicolumn{2}{c|}{\multirow{2}{*}{Methods}} & \multicolumn{3}{c|}{View-point} & \multicolumn{3}{c|}{Scale} & \multicolumn{4}{c|}{Truncation} & \multicolumn{4}{c|}{Occlusion} & \multirow{2}{*}{All} \\
             \multicolumn{2}{c|}{ }  & easy & medi & hard & easy & medi & hard & none & src & tgt & both & none & src & tgt & both & \\
            \hline
            \hline
            \multicolumn{2}{c|}{Identity mapping} & 7.3 & 3.7 & 2.6 & 7.0 & 4.3 & 3.3 & 6.5 & 4.8 & 3.5 & 5.0 & 6.1 & 4.0 & 5.1 & 4.6 & 5.6 \\
            \hline
            \multirow{6}{*}{\shortstack[1]{Authors' \\original\\ models}}
            & CNNGeo$_\textrm{res101}$~\cite{Rocco17}                & 25.2 & 10.7 & 5.9 & 22.3 & 16.1 & 8.5 & 21.1 & 12.7 & 15.6 & 13.9 & 20.0 & 14.9 & 14.3 & 12.4 & 18.1 \\
            & A2Net$_\textrm{res101}$~\cite{paul2018attentive}       & 27.5 & 12.4 & 6.9 & 24.1 & 18.5 & 10.3 & 22.9 & 15.2 & 17.6 & 15.7 & 22.3 & 16.5 & 15.2 & 14.5 & 20.1 \\ 
            & WeakAlign$_\textrm{res101}$~\cite{Rocco18}             & 29.4 & 12.2 & 6.9 & 25.4 & 19.4 & 10.3 & 24.1 & 16.0 & 18.5 & 15.7 & 23.4 & 16.7 & 16.7 & 14.8 & 21.1  \\ 
            & NC-Net$_\textrm{res101}$~\cite{rocco2018neighbourhood} & \underline{34.0} & \underline{18.6} & \underline{12.8} & \underline{31.7} & \underline{23.8} & \underline{14.2} & \underline{29.1} & \underline{22.9} & \underline{23.4} & \underline{21.0} & \underline{29.0} & \underline{21.1} & \underline{21.8} & \underline{19.6} & \underline{26.4}  \\ 
            
            
           & HPF$_{\mathrm{res50}}$~\cite{min2019hyperpixel} & \underline{35.0} & \underline{18.9} & \underline{13.6} & \underline{32.0} & \underline{25.1} & \underline{15.4} & \underline{29.7} & \underline{24.5} & \underline{23.5} & \underline{22.9} & \underline{29.6} & \underline{22.9} & \underline{22.1} & \underline{21.3} & \underline{27.2} \\
            & HPF$_{\mathrm{res101}}$~\cite{min2019hyperpixel} & \textbf{35.6} & \textbf{20.3} & \textbf{15.5} & \textbf{33.0} & \textbf{26.1} & \textbf{15.8} & \textbf{31.0} & \textbf{24.6} & \textbf{24.0} & \textbf{23.7} & \textbf{30.8} & \textbf{23.5} & \textbf{22.8} &  \textbf{21.8} & \textbf{28.2}   \\
 
            \hline
            \multirow{4}{*}{\shortstack[1]{SPair-71k \\ finetuned \\ models}}
            & CNNGeo$_\textrm{res101}$~\cite{Rocco17}                & 28.8 & 12.0 & 6.4 & 24.8 & 18.7 & 10.6 & 23.7 & 15.5 & 17.9 & 15.3 & 22.9 & 16.1 & 16.4 & 14.4 & 20.6 \\
            & A2Net$_\textrm{res101}$~\cite{paul2018attentive}       & 30.9 & 13.3 & 7.4 & 26.1 & 21.1 & 12.4 & 25.0 & 17.4 & 20.5 & 17.6 & 24.6 & 18.6 & 17.2 & 16.4 & 22.3 \\ 
            & WeakAlign$_\textrm{res101}$~\cite{Rocco18}             & 29.3 & 11.9 & 7.0 & 25.1 & 19.1 & 11.0 & 24.0 & 15.8 & 18.4 & 15.6 & 23.3 & 16.1 & 16.4 & 15.7 & 20.9 \\ 
            & NC-Net$_\textrm{res101}$~\cite{rocco2018neighbourhood} & 26.1 & 13.5 & 10.1 & 24.7 & 17.5 & 9.9 & 22.2 & 17.1 & 17.5 & 16.8 & 22.0 & 16.3 & 16.3 & 15.2 & 20.1 \\ 
            \hline
            \end{tabular}}
        \vspace{+2.0mm}
        \caption{\label{tab:HPFanalysesTable}PCK analysis by variation factors on SPair-71k. The variation factors include view-point, scale, truncation, and occlusion.} 
        \vspace{-3.0mm}
        \end{center}
    \end{table*}
\end{center}

\section{Dataset generation and annotation}

We have created the SPair-71k dataset using 1,800 images from 18 categories of PASCAL VOC~\cite{everingham2015pascal}. Specifically, we extract 1,000 images from 10 rigid categories of PASCAL 3D+~\cite{Xiang2014BeyondPA} (aeroplane, bike, boat, bottle, bus, car, chair, motorbike, train, tv/monitor)    
and 800 images of 8 non-rigid categories of PASCAL VOC 2012~\cite{everingham2015pascal} (bird, cat, cow, dog, horse, person, potted plant, sheep).  
Note that we do not use `dining table' and `sofa' categories present in PASCAL VOC as they usually appear as background and their semantic keypoints are too ambiguous to localize properly.
The images are selected to cover diverse viewpoints of each category as much as possible. 
For the selected 1,800 images, we manually annotated keypoints and generate 70,958 pairs of images with pair-level annotations as follows\footnote{When annotating keypoints, we treat `bottle', `potted plant', `train' and `tv/monitor' as flat instances as it is hard to discriminate between front/back and left/right of the instances due to their cylindrical shapes.}.


\begin{figure}[t]
	
    \centering
    \includegraphics[width=1.0\linewidth]{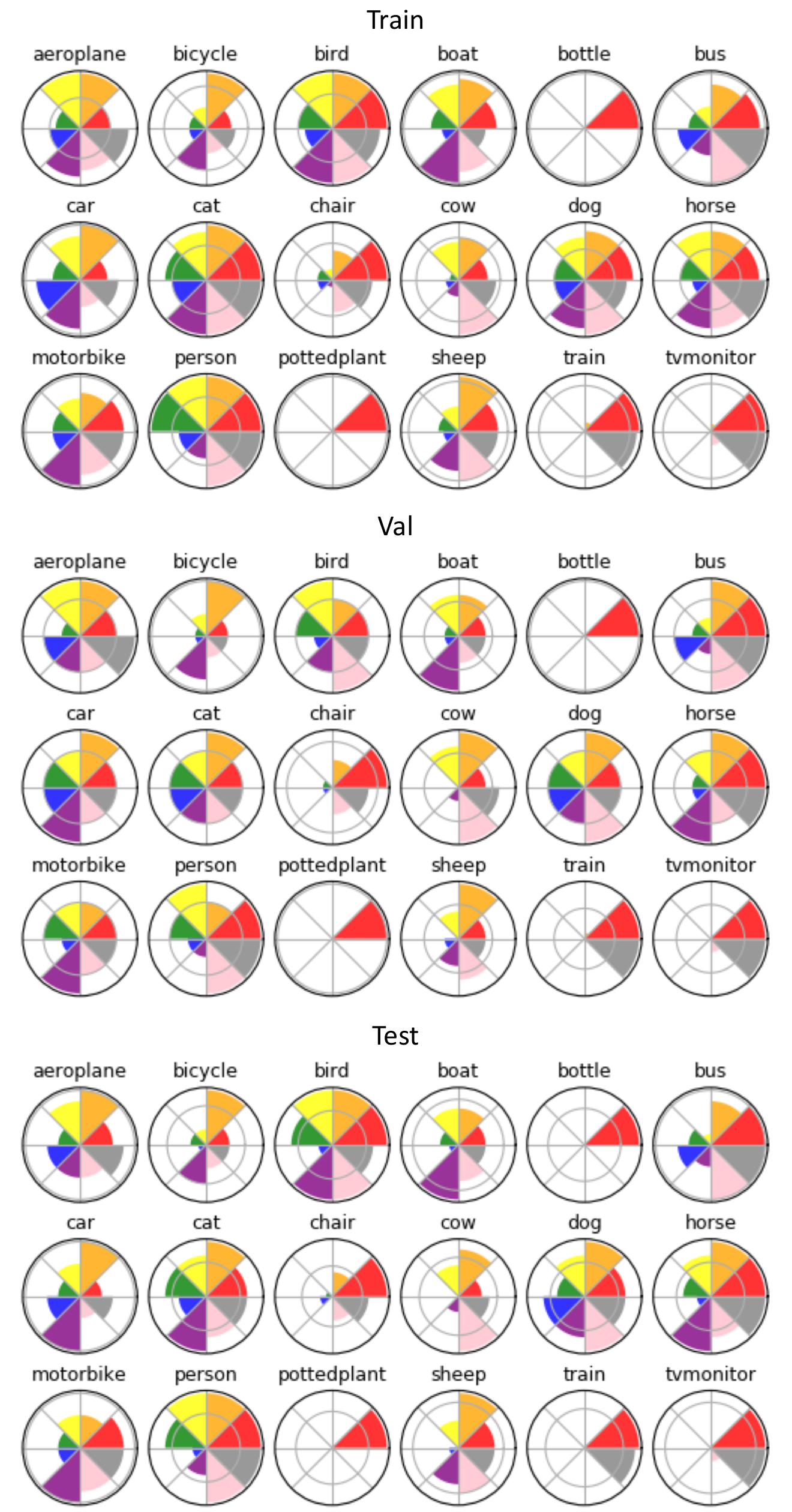}
	\vspace{-2.0mm}
		
	\caption{Azimuth distributions of SPair-71k training (top), validation (middle), and testing (bottom) images.} \label{fig:azimuth_dists}
 	\vspace{-3.0mm}
\end{figure}

\smallbreak
\noindent \textbf{Image-level annotations.}  Keypoints for each object category are carefully selected and annotated according to three keypoint selection criteria: (1) each keypoint should describe an object's part shared across instances of the same object category, (2) keypoints of an object category should be distinct from each other, and (3) keypoints of an object category should spread over the whole object. The number of selected keypoints varies from 9 (potted plant) to 30 (car) across categories; occluded or truncated keypoints are not annotated, thus varying from 3 to 30 across instances in practice. Azimuths for 10 rigid categories are directly obtained from PASCAL 3D+~\cite{Xiang2014BeyondPA} and quantized to one of the eight angular bins while azimuth bins for the other 8 non-rigid categories are manually annotated. Bounding box, segmentation mask, truncation and occlusion labels are retrieved from PASCAL VOC 2012~\cite{everingham2015pascal} where both truncation and occlusion labels are binary indicators, \ie, whether an instance in the image is truncated (occluded) or not.

\smallbreak
\noindent \textbf{Image-level splits.}
In order to build disjoint splits of image pairs for training, validation, and testing, we first create corresponding splits of images before generating pairs.
100 images of each object category are divided into three splits with an approximate ratio of 5:2:3 so that the images of each split spreads over quantized azimuth values, thus obtaining 997, 322, and 481 images for training, validation, and testing splits, respectively.
Figure~\ref{fig:azimuth_dists} shows category-wise circular histograms of azimuth values for the splits. 
Images of each split are then used to generate image pairs with pair-level annotations as follows.

\smallbreak
\noindent \textbf{Pair-level annotations.} We build pair-level annotations using keypoints, azimuth bins, bounding boxes, truncations, and occlusions annotated in images. Common keypoints in two images are used as keypoints of pair-level annotations, \ie, keypoint correspondences. If there are no common keypoints between the two, the pair is excluded.
View-point differences are divided into three levels of `easy', `medium', and `hard'; a pair is marked as `easy', `medium', and `hard' if the difference between azimuth bin indexes of the two instances falls in the range of \{0, 1\}, \{2, 3\}, and \{4\}, respectively. Scale differences are labeled levels of `easy', `medium', and `hard'; a pair is labeled `easy', `medium', and `hard' if the area ratio of corresponding object bounding boxes falls in the range of [1,2), [2, 4), [4, $\infty$]. Each pair is also annotated for both truncation and occlusion levels with `none', `source only', `target only', and `both'. 
See Table~\ref{tab:large_split_additional} and~\ref{tab:large_split} for details.

Finally, we obtain the SPair-71k dataset of 70,958 image pairs in total, which consists of 53,340 for training, 5,384 for validation, and 12,234 for testing, respectively.

\begin{center}
    \begin{table*}
        \begin{center}
            \scalebox{0.81}{
            \begin{tabular}{l|l|ccc|ccc|cccc|cccc|c}
                \hline
                \multirow{2}{*}{Approach} & \multirow{2}{*}{Methods} & \multicolumn{3}{c|}{View-point}  & \multicolumn{3}{c|}{Scale} & \multicolumn{4}{c|}{Truncation} & \multicolumn{4}{c|}{Occlusion} & \multirow{2}{*}{All}\\
                 &  & easy & medi & hard & easy & medi & hard & none & src & tgt & both & none & src & tgt & both\\
                \hline
                \hline
                \multirow{3}{*}{\shortstack[1]{Image \\ \\ alignment}}  & CNNGeo$_\textrm{res101}$~\cite{Rocco17} & 44.3  & 15.1 & 9.9 & 44.3 & 31.0 & 16.5 & 44.3 & 28.4 & 32.6 & 26.2 & 44.3 & 31.3 & 31.6 & 20.9 & 20.6 \\
                
                 & A2Net$_\textrm{res101}$~\cite{paul2018attentive} & 45.0 & 15.8 & 10.4 & 45.0 & 34.2 & 18.6 & 45.0 & 32.5 & \underline{36.4} & 30.0 & 45.0 & 32.3 & 29.7 & 24.0 & 22.3 \\
                 
                 & WeakAlign$_\textrm{res101}$~\cite{Rocco18} & 44.8 & 15.3 & 10.2 & 44.8 & 32.0 & 17.1 & 44.8 & 29.2 & 33.9 & 26.1 & 44.8 & 31.5 & 31.1 & 24.6 & 20.9 \\
                 
                \hline
                
                \multirow{3}{*}{\shortstack[1]{Region \\ \\ matching}} & NC-Net$_\textrm{res101}$~\cite{rocco2018neighbourhood} & \underline{49.8} & \underline{23.4} & \underline{19.1} & \underline{49.8} & \underline{35.7} & \underline{19.6} & \underline{49.8} & \underline{33.7} & \textbf{36.8} & \textbf{33.8} & \underline{49.8} & \underline{35.6} & \underline{35.9} & \underline{27.4} & \underline{26.4} \\
                
                 & HPF$_{\mathrm{res50}}$~\cite{min2019hyperpixel} & \underline{50.1} & \underline{22.8} & \underline{21.1} & \underline{50.1} & \underline{35.6} & \textbf{23.0} & \underline{50.1} & \textbf{37.4} & 36.0 & \underline{33.1} & \underline{50.1} & \underline{36.7} & \underline{35.6} & \underline{28.8} & \underline{27.2} \\
                 
                 & HPF$_{\mathrm{res101}}$~\cite{min2019hyperpixel} & \textbf{51.0} & \textbf{25.2} & \textbf{23.8} & \textbf{51.0} & \textbf{37.3} & \underline{22.8} & \textbf{51.0} & \underline{36.7} & \textbf{36.8} & \underline{33.1} & \textbf{51.0} & \textbf{37.6} & \textbf{36.7} & \textbf{29.6} & \textbf{28.2} \\
                \hline
            \end{tabular}}
        \vspace{+2.0mm}
        \caption{\label{tab:SPairIndepth} PCK analysis by controlling individual variations. For each variation of view-point, scale, truncation, and occlusion, the difficulty levels of the other variations are fixed as easy (view-point and scale) and none (truncation and occlusion). In this experiment, we use author's original models for \cite{min2019hyperpixel, Rocco18, rocco2018neighbourhood} and SPair-71k-finetuned models for \cite{ Rocco17, paul2018attentive} in favor of better performance.}
        \end{center}
    \end{table*}
\end{center}




\section{Baseline results on SPair-71k}

We evaluate recent state-of-the-art methods~\cite{min2019hyperpixel, Rocco17, Rocco18, rocco2018neighbourhood, paul2018attentive} on SPair-71k to provide baseline results for further research. For each method in comparison, we run two versions of each model: an original trained model provided by the authors and a model further finetuned by ourselves using SPair-71k train/val set. The results are shown in Table~\ref{tab:hpftable}. We fail to successfully train the method of~\cite{Rocco18,rocco2018neighbourhood} on SPair-71k so that their performances drop when trained. We guess that their original learning objectives for weakly-supervised learning is fragile in presence of large view-point differences as in SPair-71k. We leave this issue for further investigation and will update the results at our benchmark page.

\smallbreak
\noindent \textbf{Analysis by variation factors.} Each image pair in SPair-71k has annotations of difficulty levels for four variation factors (\ie, view-point, scale, truncation, and occlusion) between corresponding instances of the same category; `easy', `medium', or `hard' is annotated for view-point and scale changes, while `none', `source', `target' or `both' is annotated for truncation and occlusion.
PCK analysis of the models using these annotations are summarized in Table~\ref{tab:HPFanalysesTable}. The results show that all the models perform better given pairs with less variation, and that view-point and scale changes significantly affect the performances.

\smallbreak
\noindent \textbf{Impact of individual variations.}
The results in Table~\ref{tab:HPFanalysesTable} does not clearly demonstrate an impact of each individual variation because the four types of variations co-exist in a pair and interfere with each other when evaluated. 
To measure an impact of each variation individually, we need to control the other variations to remain fixed.
To this end, we evaluate the performances of different levels of a specific variation factor while fixing the levels of the other variations as easy (view-point and scale) and none (truncation and occlusion).
For example, when evaluating the performances varying view-point levels, we only use pairs that are labeled `easy' scale, `none' truncation, and `none' occlusion. 
The results are summarized in Table~\ref{tab:SPairIndepth}. It shows that the performances of local-region-matching methods~\cite{min2019hyperpixel, rocco2018neighbourhood} is more robust to view-point variation compared to global-image-alignment methods~\cite{Rocco17, Rocco18, paul2018attentive}; the performance of the image alignment models drops more quickly than those of the region matching ones. In terms of scale changes, truncation, and occlusion, however, we find no significant difference in performance drop between the methods. While both truncation and occlusion clearly degrade the performances, the impacts are less than view-point and scale variations. 

\section{Conclusion}
In this paper, we have presented a large-scale benchmark dataset, SPair-71k, which consists of 71k image pairs for semantic correspondence. 
Compared to previous datasets, it contains a significantly large number of image pairs with diverse variations in view-point, scale, truncation and occlusion, thus generalizing the problem of visual correspondence by reflecting real-world scenarios. Moreover, its rich annotations including object bounding boxes, keypoint correspondences, variation factors, azimuths, and object segmentation masks will be useful for future research on semantic correspondence and its joint problems. 

\smallbreak
\noindent \textbf{Acknowledgements.}
This work is supported by Samsung Advanced Institute of Technology (SAIT) and Basic Science Research Program (NRF-2017R1E1A1A01077999), and also in part by the Inria/NYU collaboration and the Louis Vuitton/ENS chair on artificial intelligence.

{\small
\bibliographystyle{ieee_fullname}
\bibliography{egbib}
}




\end{document}